\documentclass[10pt,twocolumn,letterpaper]{article}

\usepackage{cvpr}
\usepackage{times}
\usepackage{epsfig}
\usepackage{graphicx}
\usepackage{amsmath}
\usepackage{amssymb}
\usepackage{multirow}
\usepackage{tabularx}
\usepackage{booktabs}
\usepackage{appendix}

\usepackage[pagebackref=true,breaklinks=true,colorlinks,bookmarks=false]{hyperref}
\newcommand{\st}[1]{\textcolor{red}{\textbf{#1}}}
\newcommand{\nd}[1]{\textcolor{blue}{\textbf{#1}}}
\newcommand{\tabincell}[2]{\begin{tabular}{@{}#1@{}}#2\end{tabular}}  
\def\eqnvspace{{\vspace{-3mm}}}
\cvprfinalcopy 

\def\cvprPaperID{} 

\ifcvprfinal\fi
\begin{document}

\title{Learning Depth-Guided Convolutions for Monocular 3D Object Detection}

\author{
  Mingyu Ding\textsuperscript{\rm 1,2}~~ 
  Yuqi Huo \textsuperscript{\rm 2,5}~~
  Hongwei Yi \textsuperscript{\rm 3}~~
  Zhe Wang\textsuperscript{\rm 4}~~
  Jianping Shi\textsuperscript{\rm 4}~~ 
  Zhiwu Lu\textsuperscript{\rm 2,5}~~ 
  Ping Luo\textsuperscript{\rm 1} \\
 \textsuperscript{\rm 1}The University of Hong Kong~~
 \textsuperscript{\rm 2}Gaoling School of Artificial Intelligence, Renmin University of China~~\\
 \textsuperscript{\rm 3}Shenzhen Graduate School, Peking University~~~~~~~~~~~
 \textsuperscript{\rm 4}SenseTime Research\\
 \textsuperscript{\rm 5}Beijing Key Laboratory of Big Data Management and Analysis Methods, Beijing 100872, China\\
  \texttt{\{myding, pluo\}@cs.hku.hk}~~~
  \texttt{\{bohony,luzhiwu\}@ruc.edu.cn}~~~\\
  \texttt{hongweiyi@pku.edu.cn}~~~
  \texttt{\{wangzhe,shijianping\}@sensetime.com}
  \\
}


\maketitle

\begin{abstract}
3D object detection from a single image without LiDAR is a challenging task due to the lack of accurate depth information. Conventional 2D convolutions are unsuitable for this task because they fail to capture local object and its scale information, which are vital for 3D object detection. To better represent 3D structure, prior arts typically transform depth maps estimated from 2D images into a pseudo-LiDAR representation, and then apply existing 3D point-cloud based object detectors. However, their results depend heavily on the accuracy of the estimated depth maps, resulting in suboptimal performance. In this work, instead of using pseudo-LiDAR representation, we improve the fundamental 2D fully convolutions by proposing a new local convolutional network (LCN), termed Depth-guided Dynamic-Depthwise-Dilated LCN (D$^4$LCN), where the filters and their receptive fields can be automatically learned from image-based depth maps, making different pixels of different images have different filters. D$^4$LCN overcomes the limitation of conventional 2D convolutions and narrows the gap between image representation and 3D point cloud representation. Extensive experiments show that D$^4$LCN outperforms existing works by large margins. For example, the relative improvement of D$^4$LCN against the state-of-the-art on KITTI is 9.1\% in the moderate setting. D$^4$LCN ranks $1^{\mathrm{st}}$ on KITTI monocular 3D object detection benchmark at the time of submission (car, December 2019) . The code is available at \href{https://github.com/dingmyu/D4LCN}{https://github.com/dingmyu/D4LCN}.
\end{abstract}

\begin{figure}[t]
    \begin{center}
    \includegraphics[width=0.9\columnwidth]{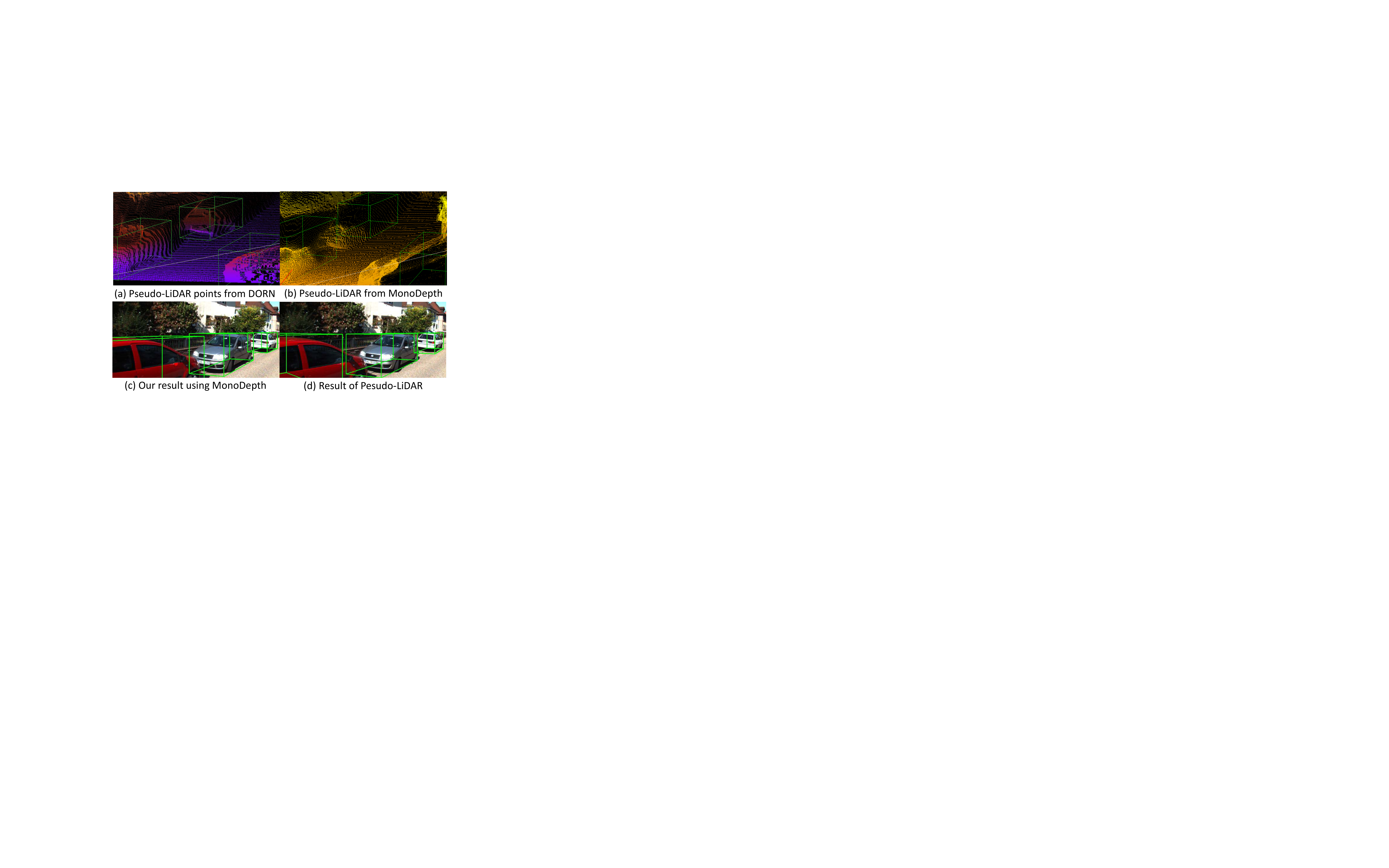}
    \end{center}
    \vspace{-10pt}
    \caption{(a) and (b) show pseudo-LiDAR points generated by the supervised depth estimator, DORN \cite{fu2018deep} and the unsupervised Monodepth \cite{monodepth17} respectively. The green box represents ground-truth (GT) 3D box. Pseudo-LiDAR points generated by inaccurate depth as shown in (b) have large offsets comapred to the GT box. (c) and (d) show the detection results of our method and Pseudo-Lidar \cite{wang2019pseudo} by using a coarse depth map. The performance of \cite{wang2019pseudo} depends heavily on the accuracy of the estimated depth maps, while our method achieves accurate detection results when accurate depth maps are missing. }
    \label{fig:first_fig}
    \vspace{-16pt}
\end{figure}

\section{Introduction}

3D object detection is a fundamental problem and has many applications such as autonomous driving and robotics. Previous methods show promising results by utilizing LiDAR device, which produces precise depth information in terms of 3D point clouds. However, due to the high-cost and sparse output of LiDAR, it is desirable to seek cheaper alternatives like monocular cameras. This problem remains largely unsolved, though it has drawn much attention. 

Recent methods towards the above goal can be generally categorized into two streams as image-based approaches \cite{mousavian20173d, li2019gs3d, qin2019monogrnet, hu2019joint, he2019mono3d++, chen2016monocular} and pseudo-LiDAR point-based approaches \cite{wang2019pseudo, ma2019accurate, weng2019monocular}. The image-based approaches~\cite{chen20153d,he2019mono3d++} typically leverage geometry constraints including object shape, ground plane, and key points. These constraints are formulated as different terms in loss function to improve detection results. The pseudo-LiDAR point-based approaches transform depth maps estimated from 2D images to point cloud representations to mimic the LiDAR signal. As shown in Figure~\ref{fig:first_fig}, both of these methods have drawbacks, resulting in suboptimal performance.

\begin{figure}[t]
    \begin{center}
    \includegraphics[width=0.97\columnwidth]{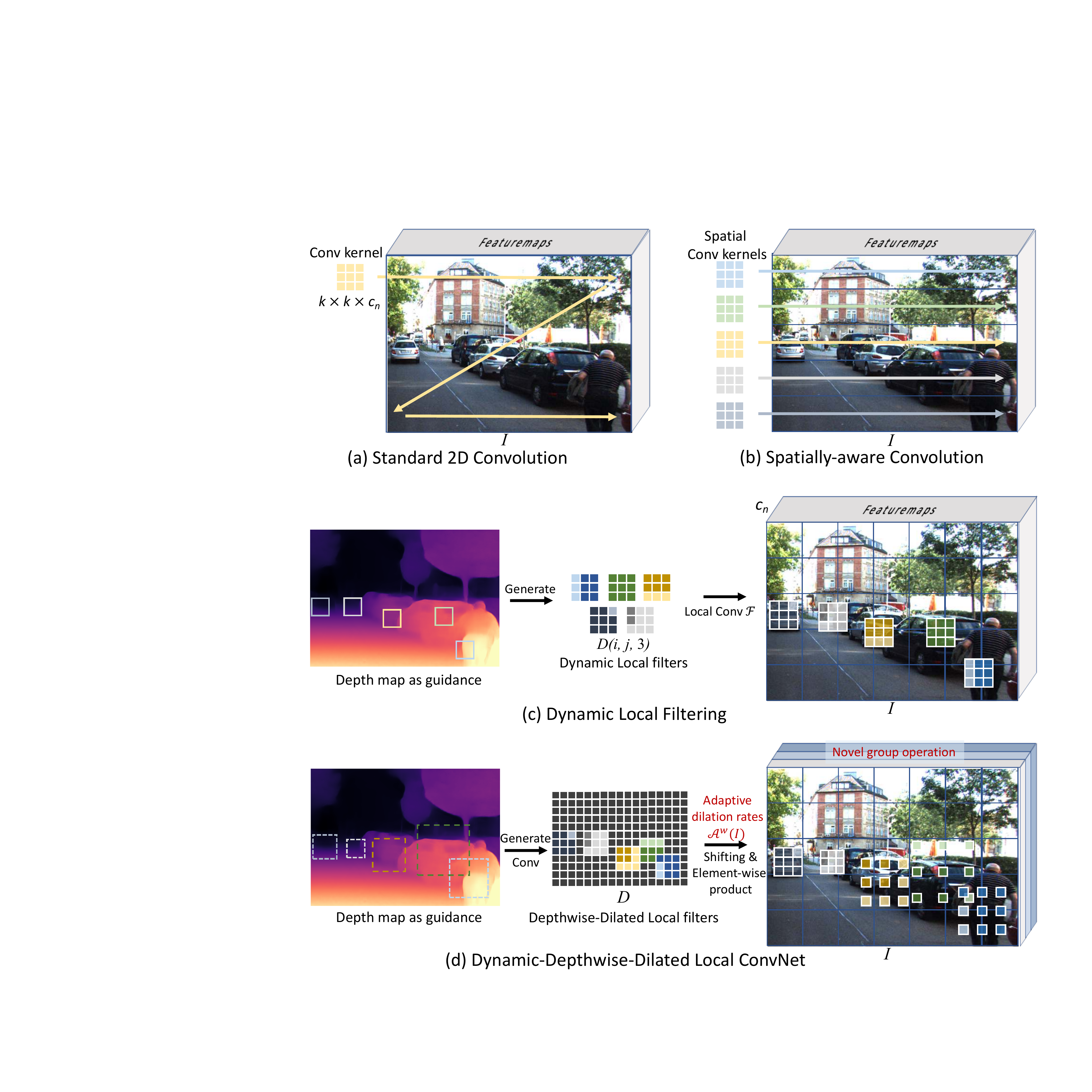}
    \end{center}
    \vspace{-10pt}
    \caption{\textbf{Comparisons} among different convolutional approaches. (a) is the traditional 2D convolution that uses a single convolutional kernel applied on each pixel to convolve the entire image. (b) applies multiple fixed convolutional kernels on different regions (slices) of an image. (c) uses the depth map to generate dynamic kernels with the same receptive fields for each pixel. (d) denotes our approach, where the filter is dynamic, depth-wise, and has adaptive receptive fields for each pixel and channel of the feature map. It can be implemented more efficiently with fewer parameters than (c). Best viewed in color. }
    \label{fig:conv}
    \vspace{-8pt}
\end{figure}

Specifically, the image-based methods typically fail to capture meaningful local object scale and structure information, because of the following two factors.
(1) Due to perspective projection, the monocular view at far and near distance would cause significant changes in object scale. It is difficult for traditional 2D convolutional kernels to process objects of different scales simultaneously (see Figure~\ref{fig:conv}). 
(2) The local neighborhood of 2D convolution is defined in the camera plane where the depth dimension is lost. In this non-metric space (\ie the distance between pixels does not have a clear physical meaning like depth), a filter cannot distinguish objects from the background. In that case, a car area and the background area would be treated equally. 

Although pseudo-LiDAR point-based approaches have achieved progressive results, they still possess two key issues. 
(1) The performance of these approaches heavily relies on the precision of estimated depth maps (see Figure~\ref{fig:first_fig}). The depth maps extracted from monocular images are often coarse (point clouds estimated using them have wrong coordinates), leading to inaccurate 3D predictions. In other words, the accuracy of the depth map limits the performance of 3D object detection. 
(2) Pseudo-LiDAR methods cannot effectively employ high-level semantic information extracted from RGB images, leading to many false alarms. This is because point clouds provide spatial information but lose semantic information. As a result, regions like roadblocks, electrical boxes and even dust on the road may cause false detection, but they can be easily discriminated by using RGB images.

To address the above problems, we propose a novel convolutional network, termed Depth-guided Dynamic-Depthwise-Dilated local convolutional network  (D$^4$LCN), where the convolutional kernels are generated from the depth map and locally applied to each pixel and channel of individual image sample, rather than learning global kernels to apply to all images. As shown in Figure~\ref{fig:conv}, D$^4$LCN treats the depth map as guidance to learn local dynamic-depthwise-dilated kernels from RGB images, so as to fill the gap between 2D and 3D representation. More specifically, the learned kernel in D$^4$LCN is sample-wise (\ie exemplar kernel \cite{he2010guided}), position-wise (\ie local convolution \cite{jia2016dynamic}), and depthwise (\ie depthwise convolution \cite{howard2017mobilenets}), where each kernel has its own dilation rate (\ie different exemplar kernels have different receptive fields). 

D$^4$LCN is carefully designed with four considerations. (1) The exemplar kernel is to learn specific scene geometry for each image. (2) The local convolution is to distinguish object and background regions for each pixel. (3) The depth-wise convolution is to learn different channel filters in a convolutional layer with different purposes and to reduce computational complexity. (4) The exemplar dilation rate is to learn different receptive fields for different filters to account for objects with diverse scales. The above delicate designs can be easily and efficiently implemented by combing linear operators of shift and element-wise product.
As a result, the efficient D$^4$LCN can not only address the problem of the scale-sensitive and meaningless local structure of 2D convolutions, but also benefit from the high-level semantic information from RGB images compared with the pseudo-LiDAR representation.

Our main \textbf{contributions} are three-fold. (1) A novel component for 3D object detection, D$^4$LCN, is proposed, where the depth map guides the learning of dynamic-depthwise-dilated local convolutions from a single monocular image. (2) We carefully design a single-stage 3D object detection framework based on D$^4$LCN to learn better 3D representation for reducing the gap between 2D convolutions and 3D point cloud-based operations. (3) Extensive experiments show that D$^4$LCN outperforms state-of-the-art monocular 3D detection methods and takes the first place on the KITTI benchmark \cite{geiger2012we}.

\section{Related Work}

\textbf{Image-based Monocular 3D Detection.}~~Previous monocular 3D detection methods \cite{mousavian20173d, li2019gs3d, qin2019monogrnet, brazil2019m3d, hu2019joint, he2019mono3d++, chen2016monocular, xu2018multi} usually make assumptions about the scene geometry and use this as a constraint to train the 2D-to-3D mapping. Deep3DBox~\cite{mousavian20173d} uses the camera matrix to project a predicted 3D box onto the 2D image plane, constraining each side of the 2d detection box, such that it corresponds to any of the eight corners of the 3D box. OFTNet~\cite{roddick2018orthographic} introduces the orthographic feature transform, which maps image-based features into an orthographic 3D space. It is helpful when scale of objects varies drastically.
\cite{jorgensen2019monocular, liu2019deep} investigated different ways of learning the confidence to model heteroscedastic uncertainty by using a 3D intersection-over-union (IoU) loss.
To introduce more prior information, \cite{chabot2017deep, kundu20183d, zeeshan2014cars, xiang2015data} used 3D shapes as templates to get better object geometry. \cite{ku2019monocular} predicts a point cloud in an object-centered coordinate system and devises a projection alignment loss to learn local scale and shape information. \cite{manhardt2019roi} proposes a 3D synthetic data augmentation algorithm via in-painting recovered meshes directly onto the 2D scenes.

However, as it is not easy for 2D image features to represent 3D structures, the above geometric constraints fail to restore accurate 3D information of objects from just a single monocular image. Therefore, our motivation is to utilize depth information, which essentially bridges gap between 2D and 3D representation, to guide learning the 2D-to-3D feature representation.

\textbf{Point Cloud-based Monocular 3D Detection.}~~Previous monocular methods \cite{wang2019pseudo, ma2019accurate, weng2019monocular} convert image-based depth maps to pseudo-LiDAR representations for mimicking the LiDAR signal. With this representation, existing LiDAR-based detection algorithms can be directly applied to monocular 3D object detection. For example, \cite{weng2019monocular} detects 2D object proposals in the input image and extracts a point cloud frustum from the pseudo-LiDAR for each proposal. \cite{ma2019accurate} proposes a multi-modal features fusion module to embed the complementary RGB cue into the generated point clouds representation.
However, this depth-to-LiDAR transformation relies heavily on the accuracy of depth map and cannot make use of RGB information.
In contrast, our method treats  depth map as guidance to learn better 3D representation from RGB images.


\textbf{LiDAR-based 3D Detection.}~~With the development of deep learning on point sets, 3D feature learning \cite{qi2017pointnet, qi2017pointnet++, zhou2018voxelnet} is able to learn deep point-based and voxel-based features. Benefit from this, LiDAR-based methods have achieved promising results in 3D detection. For example, \cite{zhou2018voxelnet} divides point clouds into equally spaced 3D voxels and transforms a group of points within each voxel into a unified feature representation.
\cite{wang2019voxel} applies the FPN technique to voxel-based detectors. 
\cite{yan2018second} investigates a sparse convolution for voxel-based networks. \cite{lang2019pointpillars} utilizes PointNets to learn a representation of point clouds organized in vertical columns (pillars).
\cite{qi2018frustum} leverages mature 2D object detectors to learn directly from 3D point clouds. \cite{wang2019frustum} aggregates point-wise features as frustum-level feature vectors.
\cite{shi2019pointrcnn, chen2019fast} directly generated a small number of high-quality 3D proposals from point clouds via segmenting the point clouds of the whole scene into foreground and background.
There are also some works focus on multi-sensor fusion (LIDAR as well as cameras) for 3D object detection.
\cite{liang2018deep, liang2019multi} proposed a continuous fusion layer that encodes both discrete-state image features as well as continuous geometric information. \cite{chen2017multi, ku2018joint} used LIDAR point clouds and RGB images to generate features and encoded the sparse 3D point cloud with a compact multi-view representation.

\begin{figure*}[t]
\centering
\includegraphics[width=0.91\textwidth]{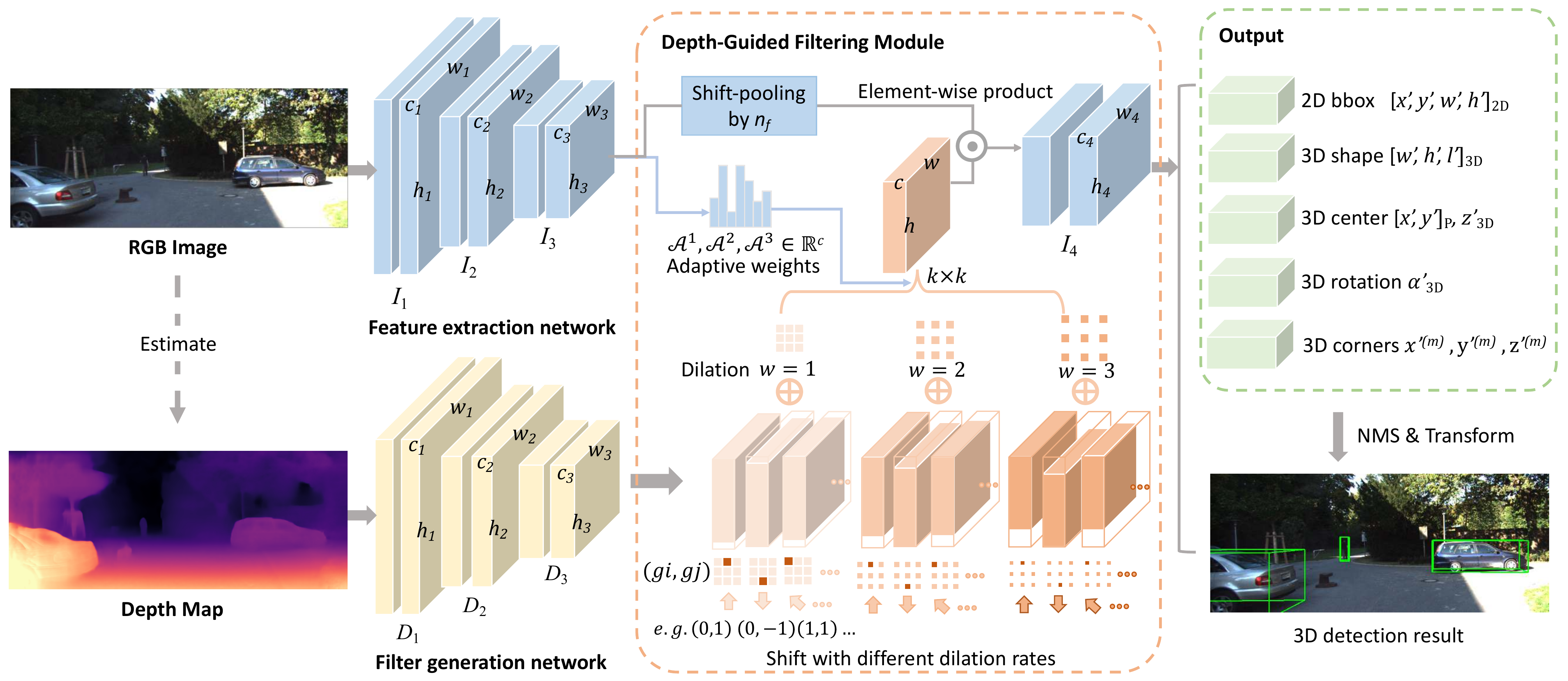}
\caption{Overview of our framework for monocular 3D object detection. The depth map is first estimated from the RGB image and used as the input of out two-branch network together with the RGB image. Then the depth-guided filtering module is used to fuse there two information of each residual block. Finally, a one-stage detection head with Non-Maximum Suppression (NMS) is employed for prediction.}
\label{fig:overview}
\vspace{-7pt}
\end{figure*}

\textbf{Dynamic Networks.}~~A number of existing techniques can be deployed to exploit the depth information for monocular 3D detection. M3D-RPN \cite{brazil2019m3d} proposes depth-aware convolution which uses non-shared kernels in the row-space to learn spatially-aware features. However, this rough and fixed spatial division has bias and fail to capture object scale and local structure.
Dynamic filtering network \cite{jia2016dynamic} uses the sample-specific and position-specific filters but has heavy computational cost, and it also fails to solve the scale-sensitive problem of 2D convolutions.
Trident network \cite{li2019scale} utilizes manually defined multi-head detectors for 2D detection. However, it needs to manually group data for different heads.
Other techniques like deformable convolution \cite{dai2017deformable} and variants of \cite{jia2016dynamic} such as \cite{ha2016hypernetworks, tang2019learning, wu2018dynamic}, fail to capture object scale and local structure
as well. In this work, our depth-guided dynamic dilated local convolutional network is proposed to solve the two problems associated with 2D convolutions and narrow the gap between 2D convolution and point cloud-based 3D processing.

\section{Methodology}

As a single-stage 3D detector, our framework consists of three key components: a network backbone, a depth-guided filtering module, and a 2D-3D detection head (see Figure~\ref{fig:overview}). Details of each component are given below. First, we give an overview of our architecture as well as backbone networks. We then detail our depth-guided filtering module which is the key component for bridging 2D convolutions and the point cloud-based 3D processing. Finally, we outline the details of our 2D-3D detection head.

\subsection{Backbone}

To utilize depth maps as guidance of 2D convolutions, we formulate our backbone as a two-branch network: the first branch is the feature extraction network using RGB images, and the other is the filter generation network to generate convolutional kernels for feature extraction network using the estimated depth as input. These two networks process the two inputs separately and their outputs of each block are merged by the depth-guided filtering module.

The backbone of the feature extraction network is ResNet-50 \cite{he2016deep} without its final FC and pooling layers, and is pre-trained on the ImageNet classification dataset \cite{deng2009imagenet}. To obtain a larger field-of-view and keep the network stride at 16, we find the last convolutional layer (conv5\_1, block4) that decreases resolution and set its stride to 1 to avoid signal decimation, and replace all subsequent convolutional layers with dilated convolutional layers (the dilation rate is 2). For the filter generation network, we only use the first three blocks of ResNet-50 to reduce computational costs. Note the two branches have the same number of channels of each block for the depth guided filtering module. 

\subsection{Depth-Guided Filtering Module}

Traditional 2D convolution kernels fail to efficiently model the depth-dependent scale variance of the objects and effectively reason about the spatial relationship between foreground and background pixels. On the other hand, pseudo-lidar representations rely too much on the accuracy of depth and lose the RGB information. To address these problems simultaneously, we propose our depth-guided filtering module. Notably, by using our module, the convolutional kernels and their receptive fields (dilation) are different for different pixels and channels of different images.

Since the kernel of our feature extraction network is trained and generated by the depth map, it is sample-specific and position-specific, as in \cite{jia2016dynamic,ha2016hypernetworks}, and thus can capture meaningful local structures as the point-based operator in point clouds.
We first introduce the idea of depth-wise convolution \cite{howard2017mobilenets} to the network, termed depth-wise local convolution (DLCN).
Generally, depth-wise convolution (DCN) involves a set of global filters, where each filter only operates at its corresponding channel, while DLCN requires a feature volume of local filters the same size as the input feature maps. As the generated filters are actually a feature volume, a naive way to perform DLCN requires to convert the feature volume into $h_n\times w_n$ location-specific filters and then apply depth-wise and local convolutions to the feature maps, where $h_n$ and $w_n$ are the height and width of the feature maps at layer $n$. This implementation would be time-consuming as it ignores the redundant computations in neighboring pixels. 
To reduce the time cost, we employ the shift and element-wise product operators, in which shift \cite{wu2018shift} is a zero-flop zero-parameter operation, and element-wise product requires little calculation. Concretely, let $I_n \in \mathbb{R}^{h_n\times w_n\times c_n}$ and $D_n \in \mathbb{R}^{h_n\times w_n\times c_n}$ be the output of the feature extraction network and filter generation network, respectively, where $n$ is the index of the block (note that block $n$ corresponds to the layer $conv_{n+1}$ in ResNet). Let $k$ denote the kernel size of the feature extraction network.
By defining a shifting grid $\{(g_i,g_j)\}, g \in (int)[1-k/2,k/2-1]$ that contains $k \cdot k$ elements, for every vector $(g_i,g_j)$, we shift the whole feature map $D$ towards the direction and step size indicated by $(g_i,g_j)$ and get the result $D^{(g_i,g_j)}$. For example, $g \in \{-1,0,1\}$ when $k=3$, and the feature map is moved towards nine directions with a horizontal or vertical step size of 0 or 1.
We then use the sum and element-wise product operations to compute our filtering result:
\begin{align}
I' = I \odot \frac{1}{k \cdot k}\sum_{g_i,g_j}D^{(g_i,g_j)}.
\label{eq:shift}
\end{align}

\begin{figure}[t]
\centering
\includegraphics[width=0.7\columnwidth]{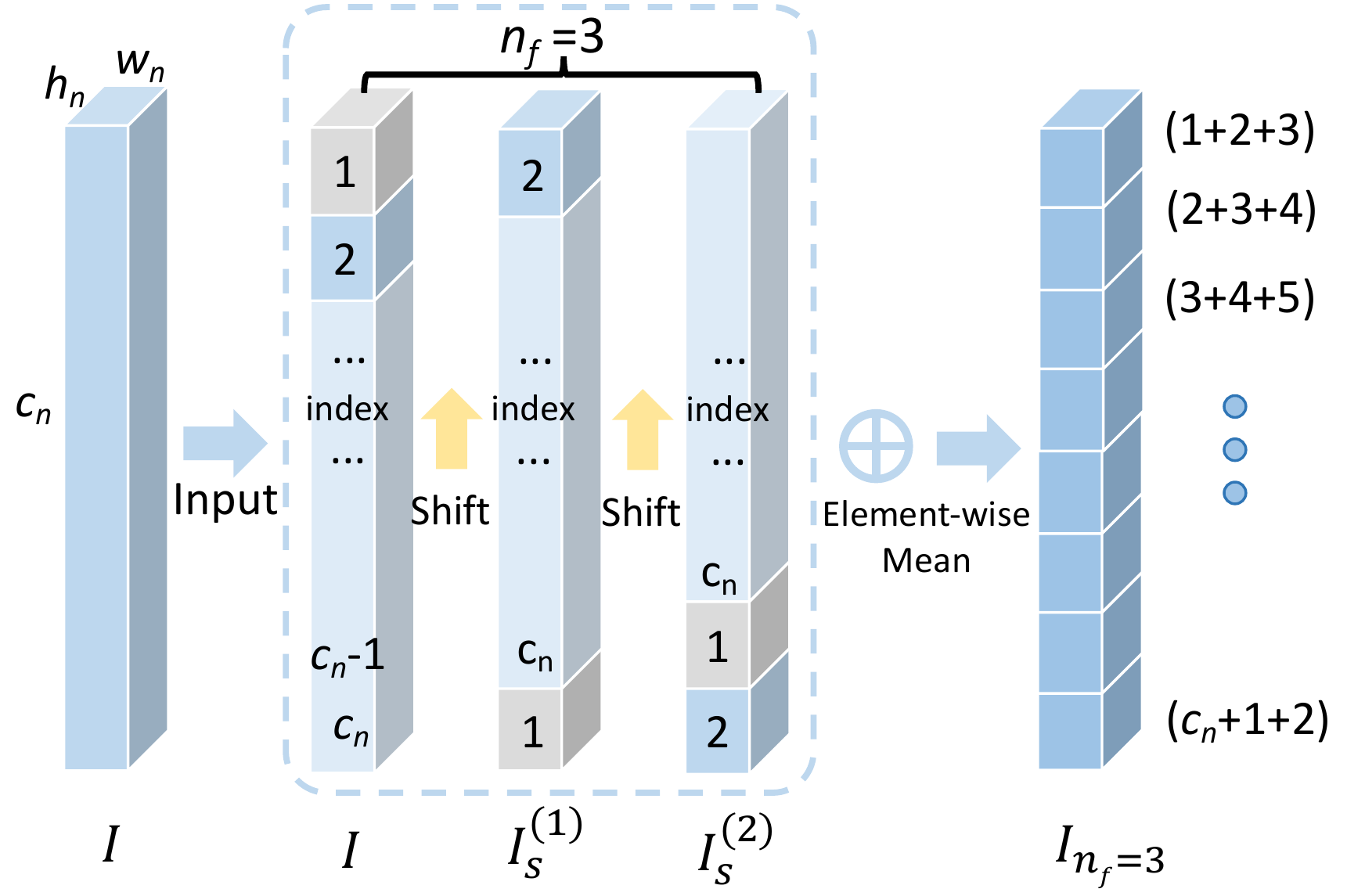}
\caption{An example of our shift-pooling operator of depth-wise convolution in depth-guided filtering module when $n_f$ is 3. It is efficiently implemented by shift and element-wise mean operators.}
\label{fig:group}
\vspace{-10pt}
\end{figure}

To encourage information flow between channels of the depth-wise convolution, we further introduce a novel shift-pooling operator in the module. Considering $n_f$ as the number of channels with information flow, we shift the feature maps along the channel axis for $n_f$ times by ${1,2,..,n_f-1}$ to obtain new $n_f-1$ shifted feature maps $I^{(n_i)}_{s},n_i \in \{1,2,...,n_f-1\}$. Then we perform element-wise mean to the shifted feature maps and the original $I$ to obtain the new feature map as the input of the module.
The process of this shift-pooling operation is shown in Figure~\ref{fig:group} ($n_f=3$). 

Compared to the idea `group' of depth-wise convolution in \cite{howard2017mobilenets,zhang2018shufflenet} which aims to group many channels into a group to perform information fusion between them, the proposed shift-pooling operator is more efficient and adds no additional parameters to the convolution. The size of our convolutional weights of each local kernel is always $k \times k \times c_n$ when applying shift-pooling, while it changes significantly in \cite{howard2017mobilenets} for different number of groups from $k \times k \times c_n$ to $k \times k \times c_n \times c_n$ in group convolution (assume that the convolution keeps the number of channels unchanged). 
Note that it is difficult for the filter generation network to generate so many kernels for the traditional convolutions $\mathcal{F}$ between all channels, and the characteristic of being position-specific dramatically increases their computational cost. 

With our depth-wise formulation, different kernels can have different functions. This enables us to assign different dilation rates \cite{yu2015multi} for each filter to address the scale-sensitive problem. Since there are huge intra-class and inter-class scale differences in an RGB image, we use $I$ to learn an adaptive dilation rate for each filter to obtain different sizes of receptive fields by an adaptive function $\mathcal{A}$. Specifically, let $d$ denote our maximum dilation rate, the adaptive function $\mathcal{A}$ consists of three layers: (1) an AdaptiveMaxPool2d layer with the output size of $d\times d$ and channel number $c$; (2) a convolutional layer with a kernel size of $d \times d$ and channel number $d \times c$; (3) a reshape and softmax layer to generate $d$ weights $\mathcal{A}^{w}(I), w \in (int)[1,d]$ with a sum of 1 for each filter. Formally, our guided filtering with adaptive dilated function (D$^4$LCN) is formulated as follows:
\begin{align}
I' = \frac{1}{d \cdot k \cdot k} \cdot I \odot \sum_w{\mathcal{A}^{w}(I)\sum_{g_i,g_j}D^{(g_i*w,g_j*w)}},
\label{eq:adaptive}
\end{align}
For different images, our depth-guided filtering module assigns different kernels on different pixels and adaptive receptive fields (dilation) on different channels. This solves the problem of scale-sensitive and meaningless local structure of 2D convolutions, and also makes full use of RGB information compared to pseudo-LiDAR representations.

\subsection{2D-3D Detection Head}

In this work, we adopt a single-stage detector with prior-based 2D-3D anchor boxes \cite{redmon2016you, liu2016ssd} as our base detector.

\subsubsection{Formulation}

\noindent \textbf{Inputs:} The output feature map $I_4 \in \mathbb{R}^{h_4 \times w_4}$ of our backbone network with a network stride factor of 16. Following common practice, we use a calibrated setting which assumes that per-image camera intrinsics $K \in \mathbb{R}^{3 \times 4}$ are available both at the training and test time. The 3D-to-2D projection can be written as:
\begin{equation}
\eqnvspace
\begin{bmatrix}
    x \\
    y \\
    1 \\
  \end{bmatrix}_{P} \cdot z_{3D} = K \cdot \begin{bmatrix}
    ~x~  \\
    ~y~  \\
    ~z~ \\
    ~1~ \\
  \end{bmatrix}_{3D}
\eqnvspace
\label{eqn:proj}
\end{equation}
where $[x,y,z]_{3D}$ denotes the horizontal position, height and depth of the 3D point in camera coordinates, and $[x,y]_P$ is the projection of the 3D point in 2D image coordinates.

\noindent \textbf{Ground Truth:} We define a ground truth (GT) box using the following parameters: the 2D bounding box $[x,y,w,h]_{2D}$, where $(x,y)$ is the center of 2D box and $w,h$ are the width and height of 2D box; the 3D center $[x,y,z]_{3D}$ represents the location of 3D center in camera coordinates; the 3D shapes $[w,h,l]_{3D}$ (3D object dimensions: height, width, length (in meters)), and the allocentric pose $\alpha_{3D}$ in 3D space (observation angle of object, ranging $[-\pi,\pi]$) \cite{manhardt2019roi}. Note that we use the minimum enclosing rectangle of the projected 3D box as our ground truth 2D bounding box. 

\noindent \textbf{Outputs:} Let $n_a$ denote the number of anchors and $n_c$ denote the number of classes. For each position $(i,j)$ of the input, the output for an anchor contains $35+n_c$ parameters: $\{[t_x,t_y,t_w,t_h]_{2D}, [t_x,t_y]_{P}, [t_z,t_w,t_h,t_l,t_{\alpha}]_{3D}$, $t_C^{(m)}, \mathbf{s}\}$, where $[t_x,t_y,t_w,t_h]_{2D}$ is the predicted 2D box; $[t_x,t_y]_{P}$ is the position of the projected 3D corner in the 2D plane, $[t_z,t_w,t_h,t_l,t_{\alpha}]_{3D}$ denotes the depth, predicted 3D shape and rotation, respectively;  $t_C^{(m)} = \{[t_x^{(m)}, t_y^{(m)}]_{P}, [t_z^{(m)}]_{3D}\}, m \in \{1,2,...,8\}$ denotes 8 projected 3D corners; $\mathbf{s}$ denotes the classification score of each class. The size of the output is $h_4 \times w_4 \times n_a \times (35+n_c)$, where $(h_4,w_4)$ is the size of the input image with a down sampling factor of 16. The output is actually an anchor-based transformation of the 2D-3D box.

\subsubsection{2D-3D Anchor}

Inspired by \cite{brazil2019m3d}, we utilize 2D-3D anchors with priors as our default anchor boxes. More specifically, a 2D-3D anchor is first defined on the 2D space as in \cite{liu2016ssd} and then use the corresponding priors in the training dataset to calculate the part of it in the 3D space. One template anchor is defined using parameters of both spaces: $\{[A_x, A_y, A_w, A_h]_{2D}, [A_z, A_w, A_h, A_l, A_{\alpha}]_{3D}\}$, where $[A_z, A_w, A_h, A_l, A_{\alpha}]_{3D}$ denotes the 3D anchor (depth, shape, rotation).

For 2D anchors $[A_x, A_y, A_w, A_h]_{2D}$, we use 12 different scales ranging from 30 to 400 pixels in height following the power function of $30 * 1.265^{exp}, exp \in (int)[0,11]$ and aspect ratios of $[0.5, 1.0, 1.5]$ to define a total of 36 anchors. We then project all ground truth 3D boxes to the 2D space. For each projected box, we calculate its intersection over union (IoU) with each 2D anchor and assign the corresponding 3D box to the anchors that have an IoU $\geq 0.5$. For each 2D anchor, we thus use the statistics across all matching ground truth 3D boxes as its corresponding 3D anchor $[A_z, A_w, A_h, A_l, A_{\alpha}]_{3D}$. Note that we use the same anchor parameters $[A_x, A_y]_{2D}$ for the regression of $[t_x,t_y]_{2D}$ and $[t_x,t_y]_{P}$. The anchors enable our network to learn a relative value (residual) of the ground truth, which significantly reduces the difficulty of learning.

\subsubsection{Data Transformation}

We combine the output of our network which is an anchor-based transformation of the 2D-3D box and the pre-defined anchors to obtain our estimated 3D boxes: 
\begin{small}
\begin{align}
& [x',y']_{2D} = [A_x,A_y]_{2D} + [t_x,t_y]_{2D} * [A_w, A_h]_{2D} \notag \\
& [x',y']_{P} = [A_x,A_y]_{2D} + [t_x,t_y]_{P} * [A_w, A_h]_{2D} \notag \\
& [x'^{(m)}, y'^{(m)}]_{P} = [A_x,A_y]_{2D} + [t_x^{(m)}, t_y^{(m)}]_{P} * [A_w, A_h]_{2D} \notag \\
& [w', h']_{2D} = [A_w,A_h]_{2D} \cdot \exp([t_w,t_h]_{2D}) \notag \\
& [w', h', l']_{3D} = [A_w,A_h,A_l]_{3D} \cdot \exp([t_w,t_h,t_l]_{3D}) \notag \\
& [z', z'^{(m)},\alpha']_{3D} = [A_z, A_z, A_{\alpha}] + [t_z, t_z, t_{alpha}]_{3D}.
\label{eq:tranform}
\end{align}
\end{small}
where $[x',y']_{P}, [z', z'^{(m)},\alpha']_{3D}$ denote respectively the estimated 3D center projection in 2D plane, the depth of 3D center and eight corners, the 3D rotation by combining output of the network and the anchor.

\subsubsection{Losses}

Our overall loss contains a classification loss, a 2D regression loss, a 3D regression loss and a 2D-3D corner loss. We use the idea of focal loss \cite{lin2017focal} to balance the samples. Let $s_t$ and $\gamma$ denote the classification score of target class and the focusing parameter, respectively. We have:
\begin{align}
& L = (1-s_t)^{\gamma}(L_{class} + L_{2d} + L_{3d} + L_{corner}),
\label{eq:loss}
\end{align}
where $\gamma = 0.5$ in all experiments, and $L_{class}$, $L_{2d}$, $L_{3d}$, $L_{corner}$ are the classification loss, 2D regression loss, 3D regression loss and D-3D corner loss, respectively.

In this work, we employ the standard cross-entropy (CE) loss for classification:
\begin{align}
L_{class} = -\log(s_t).
\label{eq:loss_classification}
\end{align}
Moreover, for both 2D and 3D regression, we simply use the SmoothL1 regression losses:
\begin{small}
\begin{align}
& L_{2D} = SmoothL1([x',y',w',h']_{2D}, [x,y,w,h]_{2D}), \notag \\
& L_{3D} = SmoothL1([w',h',l',z',\alpha']_{3D}, [w,h,l,z,\alpha]_{3D}), \notag \\
& ~~~~~~~~ + SmoothL1([x',y']_{P}, [x,y]_{P}),  \notag \\
& L_{corner} = \frac{1}{8} \sum SmoothL1([x'^{(m)}, y'^{(m)}]_{P}, [x^{(m)},y^{(m)}]_P) \notag \\
& ~~~~~~~~ + SmoothL1([z'^{(m)}]_{3D}, [z]_{3D}),
\label{eq:loss_regression}
\end{align}
\end{small}
where $[x^{(m)},y^{(m)}]_P$ denotes the projected corners in image coordinates of the GT 3D box and $[z]_{3D}$ is its GT depth.

\begin{table*}[t] 
    \centering
    {\footnotesize  \scalebox{0.98}{  \tabcolsep=0.06cm
    \begin{tabular}{l|ccc|ccc|ccc}
    \toprule
    \multirow{2}{*}{Method}& \multicolumn{3}{c|}{Test set} & \multicolumn{3}{c|}{Split1} & \multicolumn{3}{c}{Split2}\\
    & Easy &\textbf{Moderate} & Hard & Easy & \textbf{Moderate} & Hard & Easy & \textbf{Moderate} & Hard\\
	\midrule\multirow{1}{*}{OFT-Net \cite{roddick2018orthographic}}
	& 1.61 & 1.32 & 1.00 & 4.07 & 3.27 & 3.29 & -- & -- & --\\
	\multirow{1}{*}{FQNet \cite{liu2019deep}}
	& 2.77 & 1.51 & 1.01 & 5.98 & 5.50 & 4.75 & 5.45 & 5.11 & 4.45 \\
	\multirow{1}{*}{ROI-10D \cite{manhardt2019roi}}
	& 4.32 & 2.02 & 1.46 & 10.25 & 6.39 & 6.18 & -- & -- & --\\
	\multirow{1}{*}{GS3D \cite{li2019gs3d}}
	& 4.47 & 2.90 & 2.47 & 13.46 & 10.97 & 10.38 & 11.63 & 10.51 & 10.51\\
	\multirow{1}{*}{Shift R-CNN \cite{naiden2019shift}}
	& 6.88 & 3.87 & 2.83 & 13.84 & 11.29 & 11.08 &--&--&--\\
	\multirow{1}{*}{MonoGRNet \cite{qin2019monogrnet}}
	& 9.61 & 5.74 & 4.25 & 13.88 & 10.19 & 7.62 & -- & -- & --\\
	\multirow{1}{*}{MonoPSR \cite{ku2019monocular}}
	& 10.76 & 7.25 & 5.85 & 12.75 & 11.48 & 8.59 & 13.94 & 12.24 & 10.77\\
    \multirow{1}{*}{Mono3D-PLiDAR \cite{weng2019monocular}}
	& 10.76 & 7.50 & 6.10 & 31.5 & 21.00 & 17.50 & -- & -- & --\\
	\multirow{1}{*}{SS3D \cite{jorgensen2019monocular}}
	& 10.78 & 7.68 & 6.51 & 14.52 & 13.15 & 11.85 & 9.45 & 8.42 & 7.34 \\
	\multirow{1}{*}{MonoDIS \cite{simonelli2019disentangling}}
	& 10.37 & 7.94 & 6.40 & 11.06 & 7.60 & 6.37 &--&--&--\\
    \multirow{1}{*}{Pseudo-LiDAR \cite{wang2019pseudo}}
	& -- & -- & -- & 19.50 & 17.20 & 16.20 & -- & -- & --\\
	\multirow{1}{*}{M3D-RPN \cite{brazil2019m3d}}
	& 14.76 & 9.71 & 7.42 & 20.27 & 17.06 & 15.21 & \nd{20.40} & \nd{16.48} & \nd{13.34} \\
    \multirow{1}{*}{AM3D \cite{brazil2019m3d}}
	& \nd{16.50} & \nd{10.74} & \st{9.52 (+0.01)} & \st{32.23 (+5.26)} & \nd{21.09} & \nd{17.26} &--&--&--\\
	\midrule\multirow{1}{*}{\bf D$^4$LCN (Ours)}
	& \st{16.65 (+0.15)} & \st{11.72 (+0.98)} & \nd{9.51} & \nd{26.97} & \st{21.71 (+0.62)} & \st{18.22 (+0.96)} & \st{24.29 (+3.89)} & \st{19.54 (+3.06)} & \st{16.38 (+3.04)} \\
    \bottomrule
    \end{tabular}
    }}
    \caption{Comparative results on the KITTI 3D object detection dataset. For the test set, only $\text{AP}|_{R_{40}}$ is provided by the official leaderboard. We thus show the results on the test set in $\text{AP}|_{R_{40}}$ and split1/split2 in $\text{AP}|_{R_{11}}$. We use \st{red} to indicate the highest result with relative improvement in parentheses and \nd{blue} for the second-highest result of the class \emph{car}. Our method achieves 7 firsts and 2 seconds in 9 items.}
    \label{tab:testing}
    \vspace{-8pt}
\end{table*}

\vspace{-0.0cm}
\section{Experiments}

\subsection{Dataset and Setting}

\noindent \textbf{KITTI Dataset.}~~The KITTI 3D object detection dataset \cite{geiger2012we} is widely used for monocular and LiDAR-based 3D detection. It consists of 7,481 training images and 7,518 test images as well as the corresponding point clouds and the calibration parameters, comprising a total of 80,256 2D-3D labeled objects with three object classes: Car, Pedestrian, and Cyclist. Each 3D ground truth box is assigned to one out of three difficulty classes (easy, moderate, hard) according to the occlusion and truncation levels of objects. There are two train-val splits of KITTI: the split1 \cite{chen20153d} contains 3,712 training and 3,769 validation images, while the split2 \cite{xiang2015data} uses 3,682 images for training and 3,799 images for validation. The dataset includes three tasks: 2D detection, 3D detection, and Bird's eye view, among which 3D detection is the focus of 3D detection methods.

\noindent\textbf{Evaluation Metrics.}~~Precision-recall curves are used for evaluation (with the IoU threshold of 0.7). Prior to Aug. 2019, 11-point Interpolated Average Precision (AP) metric $\text{AP}|_{R_{11}}$ proposed in the Pascal VOC benchmark is separately computed on each difficulty class and each object class. After that, the 40 recall positions-based metric $\text{AP}|_{R_{40}}$ is used instead of $\text{AP}|_{R_{11}}$, following \cite{simonelli2019disentangling}. All methods are ranked by $\text{AP}|_{R_{11}}$ of the 3D car detection in the moderate setting. 

\noindent\textbf{Implementation Details.}~~We use our depth-guided filtering module three times on the first three blocks of ResNet, which have different network strides of 4,8,16, respectively. \cite{fu2018deep} is used for depth estimation. A drop-channel layer with a drop rate of 0.2 is used after each module and a dropout layer with a drop rate of 0.5 is used after the output of the network backbone. For our single-stage detector, we use two convolutional layers as our detection head. The number of channels in the first layer is 512, and $n_a * (35 + n_c)$ for the second layer, where $n_c$ is set to 4 for three object classes and the background class, and $n_a$ is set to 36. Non Maximum Suppression (NMS) with an IoU threshold of 0.4 is used on the network output in 2D space. Since the regression of the 3D rotation $\alpha$ is more difficult than other parameters, a hill-climbing post-processing step is used for optimizing $\alpha$ as in \cite{brazil2019m3d}. The input images are scaled to $512 \times 1760$ and horizontal flipping is the only data augmentation. $n_f$ is set to 2 and the maximum dilation rate $d$ is set to 3 in all experiments. 

The network is optimized by stochastic gradient descent (SGD), with a momentum of 0.9 and a weight decay of 0.0005. We take a mini-batch size of 8 on 4 Nvidia Tesla v100 GPUs (16G). We use the `poly' learning rate policy and set the base learning rate to 0.01 and power to 0.9. The iteration number for the training process is set to 40,000. 

\subsection{Comparative Results}

We conduct experiments on the official test set and two splits of validation set of the KITTI dataset. Table~\ref{tab:testing} includes the top 14 monocular methods in the leaderboard, among which our method ranks top-1. We can observe that:
(1) Our method outperforms the second-best competitor for monocular 3D car detection by a large margin (relatively 9.1\% for 10.74 vs. 11.72) under the moderate setting (which is the most important setting of KITTI).
(2) Most competitors, such as \cite{ku2019monocular, ma2019accurate, simonelli2019disentangling, naiden2019shift, weng2019monocular, brazil2019m3d}, utilize the detector (e.g. Faster-RCNN) pre-trained on COCO/KITTI or resort to multi-stage training to obtain better 2D detection and stable 3D results, while our model is trained end-to-end using the standard ImageNet pre-trained model. 
However, we still achieve the state-of-the-art 3D detection results, validating the effectiveness of our D$^4$LCN to learn 3D structure.
(3) Recently KITTI uses $\text{AP}|_{R_{40}}$ instead of $\text{AP}|_{R_{11}}$, however, all existing methods report the results under the old metric. We thus also give results under $\text{AP}|_{R_{11}}$ on the validation set for fair comparison. It can be seen that our method outperforms all others on the two splits for 3D car detection. Our results under $\text{AP}|_{R_{40}}$ on validation set are shown in ablation study.

\begin{table}[t]
    \centering
    {\footnotesize \scalebox{0.88}{ \tabcolsep=0.1cm
    \begin{tabular}{ll|ccc|ccc}
        \toprule
        \multirow{2}{*}{Method} & \multirow{2}{*}{Task} & \multicolumn{3}{c|}{$\text{AP}|_{R_{11}}$} & \multicolumn{3}{c}{$\text{AP}|_{R_{40}}$} \\
        &  & Easy & \textbf{Moderate} & Hard & Easy & \textbf{Moderate} & Hard \\
        \midrule
	\multirow{3}{*}{3DNet} & 2D detection & 93.42 & 85.16 & 68.14 & 94.13 & 84.45 & 65.73  \\
	& \textbf{3D detection} & 17.94 & 14.61 & 12.74 & 16.72 & 12.13 & 09.46 \\
	& Bird's-eye view & 24.87 & 19.89 & 16.14 & 23.19 & 16.67 & 13.39 \\
        \midrule
	\multirow{3}{*}{+CL}& 2D detection & \textbf{94.04} & \textbf{85.56} & 68.50 & \textbf{94.98} & 84.93 & 66.11  \\
	& \textbf{3D detection} & 20.66 & 15.57 & 13.41 & 17.10 & 12.09 & 09.47\\
	& Bird's-eye view & 2903 & 23.82 & 19.41 & 24.12 & 17.75 & 13.66 \\
        \midrule
	\multirow{3}{*}{+DLCN}& 2D detection & 92.98 & 85.35 & 68.63 & 93.81 & 86.71 & 70.19  \\
	& \textbf{3D detection} & 23.25 & 17.92 & 15.58 & 18.32 & 13.50 & 10.61 \\
	& Bird's-eye view & 27.76 & 22.89 & 18.73 & 26.78 & 18.68 & 15.14 \\
        \midrule
	\multirow{3}{*}{+SP}& 2D detection & 92.57 & 85.14 & 68.40 & 93.35 & 86.52 & 67.93  \\
	& \textbf{3D detection} & 25.30 & 19.02 & 17.26 & 19.69 & 14.44 & 11.52 \\
	& Bird's-eye view & 31.39 & 24.40 & 19.85 & 26.91 & 20.07 & 15.77 \\
        \midrule
	\multirow{3}{*}{\textbf{D}$\mathbf{^4}$\textbf{LCN}}& 2D detection & 93.59 & 85.51 & \textbf{68.81} & 94.25 & \textbf{86.93} & \textbf{70.34}  \\
	& \textbf{3D detection} & \textbf{26.97} & \textbf{21.71} & \textbf{18.22} & \textbf{22.32} & \textbf{16.20} & \textbf{12.30} \\
	& Bird's-eye view & \textbf{34.82} & \textbf{25.83} & \textbf{23.53} & \textbf{31.53} & \textbf{22.58} & \textbf{17.87} \\
        \bottomrule
    \end{tabular}}
    }
    \caption{Ablation study on the class \textit{car} on the KITTI split1.}
    \label{tab:ablition}
    \vspace{-10pt}
\end{table}

\subsection{Detailed Analysis}

\subsubsection{Ablation Study}

To conduct ablation study on our model, we make comparison among five versions of our model: (1) 3DNet: the baseline model using $L_{2D}$ and $L_{3D}$ without our depth-guided filtering module; (2) + CL: the Corner Loss is added to 3DNet; (3) + DLCN: depth-guided depth-wise local filtering is added; (4) + SP: shift-pooling operator is added (with $n_f=3$); (5) D$^4$LCN (our full model): adaptive dilation rates are added, as in Eq.~\ref{eq:adaptive}. From Table~\ref{tab:ablition}, we can observe that:
(1) The performance continuously increases when more components are used for 3D object detection, showing the contribution of each component. 
(2) Our depth-guided filtering module increases the 3D detection AP scores (moderate) from \{15.57, 12.09\} to \{21.71, 16.20\} w.r.t. the $\text{AP}|_{R_{11}}$ and $\text{AP}|_{R_{40}}$ metrics, respectively. This suggests that it is indeed effective to capture the meaningful local structure for 3D object detection. 
(3) The main improvement comes from our adaptive dilated convolution (2.69 and 1.76 for $\text{AP}|_{R_{11}}$ and $\text{AP}|_{R_{40}}$, respectively), which allows each channel of the feature map to have different receptive fields and thus solves the scale-sensitive problem. Note that we have tried different values of $n_f \in \{1,2,3,4,5,6\}$, and found that $n_f=3$ is the best.

\subsubsection{Evaluation of Depth Maps}

To study the impact of accuracy of depth maps on the performance of our method, we extract depth maps using four different methods \cite{monodepth17, fu2018deep, mayer2016large, chang2018pyramid} and then apply them to 3D detection. As reported in previous works on depth estimation, the three supervised methods (\ie PSMNet, DispNet, and DORN) significantly outperform the unsupervised method \cite{monodepth17}. Among the supervised methods, Stereo-based methods \cite{chang2018pyramid,mayer2016large} are better than monocular-based DORN. With these conclusions, we have the following observations from Table~\ref{tab:depth}: (1) The accuracy of 3D detection is higher with better depth map. This is because that better depth map can provide better scene geometry and local structure. (2) As the quality of depth map increases, the growth of detection accuracy becomes slower. (3) Even with the depth maps obtained by unsupervised learning \cite{monodepth17}, our method achieves state-of-the-art results. Compared to the pseudo-lidar based method \cite{ma2019accurate}, our method relies less on the quality of depth maps (19.63 vs. 15.45 using MonoDepth).

\begin{table}[t]
    \centering
    {\footnotesize \tabcolsep=0.1cm
    \begin{tabular}{l|ccc|ccc}
        \toprule
        \multirow{2}{*}{Depth} & \multicolumn{3}{c|}{$\text{AP}|_{R_{11}}$} & \multicolumn{3}{c}{$\text{AP}|_{R_{40}}$}\\
         & Easy & \textbf{Moderate} & Hard & Easy & \textbf{Moderate} & Hard \\
        \midrule
	\multirow{1}{*}{MonoDepth \cite{monodepth17}}  & 22.43 & 19.63 & 16.38 & 16.82 & 13.18 & 10.87\\
	\multirow{1}{*}{DORN \cite{fu2018deep}} & 26.97 & 21.71 & 18.22 & 22.32 & 16.20 & 12.30 \\
	\multirow{1}{*}{DispNet \cite{mayer2016large}}  & \textbf{30.95} & 24.06 & 20.29 & \textbf{25.73} & 18.56 & 15.10 \\
	\multirow{1}{*}{PSMNet \cite{chang2018pyramid}} & 30.03 & \textbf{25.41} & \textbf{21.63} & 25.24 & \textbf{19.80} & \textbf{16.45} \\
        \bottomrule
    \end{tabular}}
    \caption{Comparisons of depth maps of different quality for 3D detection on the class \textit{car} on the KITTI split1.}
    \label{tab:depth}
    \vspace{-0pt}
\end{table}

\begin{table}[t]
    \centering
    {\footnotesize \tabcolsep=0.07cm
    \begin{tabular}{l|ccc|ccc}
        \toprule
        \multirow{2}{*}{Conv module} & \multicolumn{3}{c|}{$\text{AP}|_{R_{11}}$} & \multicolumn{3}{c}{$\text{AP}|_{R_{40}}$}\\
         & Easy & \textbf{Moderate} & Hard & Easy & \textbf{Moderate} & Hard \\
        \midrule
	Dynamic \cite{jia2016dynamic}& 23.01 & 17.67 & 15.85 & 17.47 & 12.18 & 09.53 \\
	Dynamic Local \cite{jia2016dynamic}& 25.15 & 18.42 & 16.27 & 21.09 & 13.93 & 11.31 \\
	Deformable \cite{dai2017deformable}& 23.98 & 18.24 & 16.11 & 19.05 & 13.42 & 10.07 \\
	\textbf{D}$\mathbf{^4}$\textbf{LCN (ours)} & \textbf{26.97} & \textbf{21.71} & \textbf{18.22} & \textbf{22.32} & \textbf{16.20} & \textbf{12.30} \\
        \bottomrule
    \end{tabular}}
    \caption{Comparisons of different convolutional modules for \textit{car} 3D detection on the KITTI split1.}
    \label{tab:conv}
    \vspace{-5pt}
\end{table}

\subsubsection{Evaluation of Convolutional Appoaches}

To show the effectiveness of our guided filtering module for 3D object detection, we compare it with several alternatives: Dynamic Convolution \cite{jia2016dynamic}, Dynamic Local Filtering \cite{jia2016dynamic}, and Deformable Convolution \cite{dai2017deformable}. Our method belongs to dynamic networks but yields less computation cost and stronger representation. For the first two methods, we conduct experiments using the same depth map as ours. For the third method, we apply deformable convolution on both RGB and depth branches and merge them by element-wise product. From Table~\ref{tab:conv}, we can observe that our method performs the best. This indicates that our method can better capture 3D information from RGB images due to the special design of our D$^4$LCN.

\begin{table}[t] 
    \centering
    {\footnotesize 
    \scalebox{1}{ 
    \tabcolsep=0.12cm
    \begin{tabular}{l|c|c|c}
        \toprule
        \multirow{2}{*}{Class} & Easy & \textbf{Moderate} & Hard \\
        & [split1/split2/test] & [split1/split2/test] & [split1/split2/test] \\
	\midrule\multirow{1}{*}{Car} & {26.97/24.29/16.65} & {21.71/19.54/11.72} & {18.22/16.38/9.51} \\
	Pedestrian & 12.95/12.52/4.55 & 11.23/10.37/3.42 & 11.05/10.23/2.83 \\
	Cyclist & 5.85/7.05/2.45 & 4.41/6.54/1.67 & 4.14/6.54/1.36 \\
    \bottomrule
    \end{tabular}
    }}
    \caption{Multi-class 3D detection results of our method on the three data splits. Note that all pseudo-LiDAR based methods \cite{ma2019accurate,weng2019monocular,wang2019pseudo} fail to detect pedestrians and cyclists.}
    \label{tab:multiclass}
\end{table}

\begin{figure}[t]
\centering
\includegraphics[width=0.97\columnwidth]{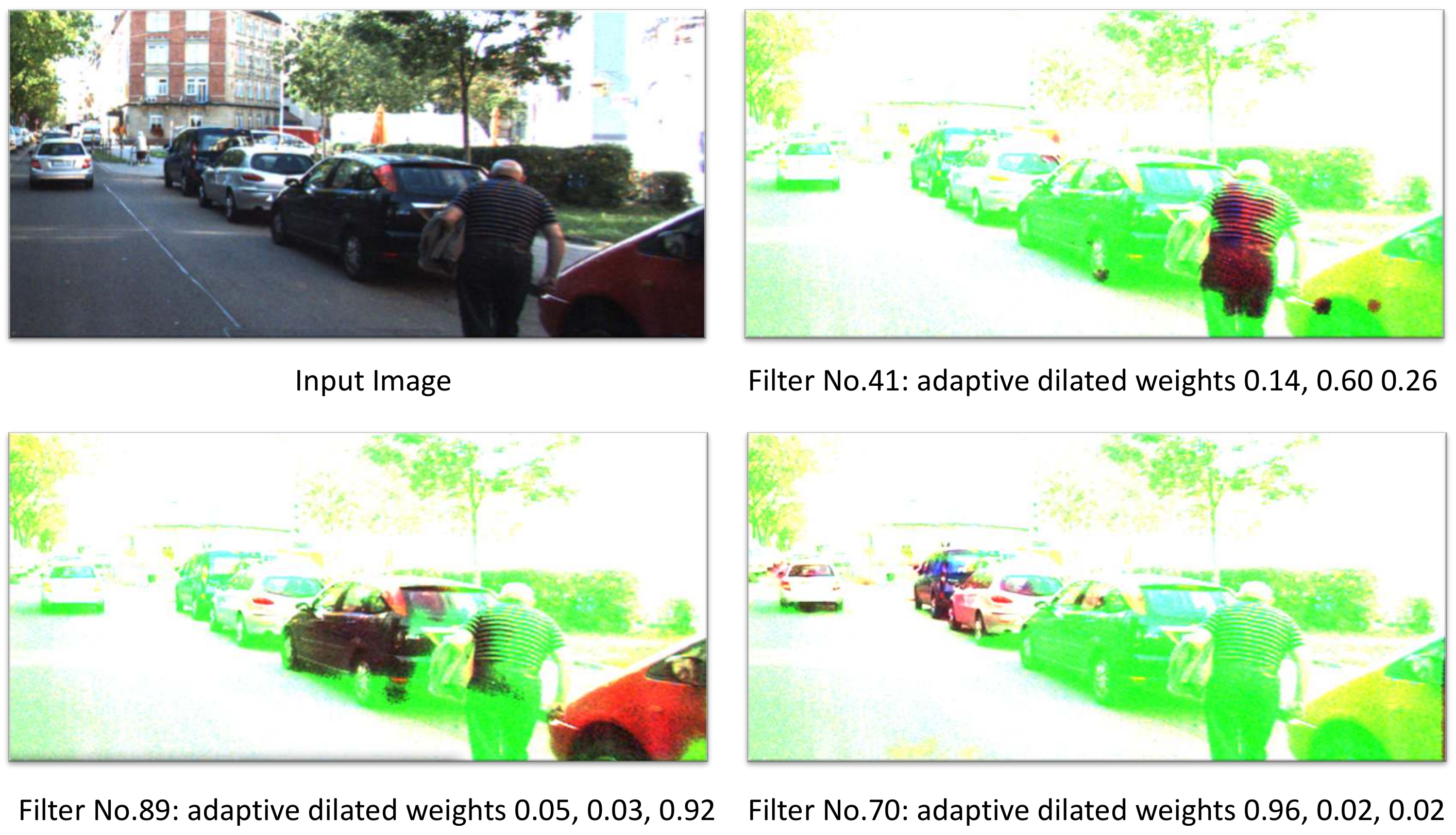}
   \caption{Visualization of active maps corresponding to different filters of block 3 of our D$^4$LCN. Each filter learns three weights representing dilation rate of 1, 2, 3, respectively. Different filters have different functions in our model to handle the scale problem adaptively. For example, filter 89 has large receptive fields for large-scale cars, while filter 70 deals with the small-scale cars.}
\label{fig:filter}
\vspace{-5pt}
\end{figure}

\subsubsection{Multi-Class 3D Detection}

Since a person is a non-rigid body, its shape varies and its depth information is hard to accurately estimate. For this reason, 3D detection of pedestrians and cyclists becomes particularly difficult. Note that all pseudo-LiDAR based methods \cite{ma2019accurate, weng2019monocular, wang2019pseudo} fail to detect these two categories. However, as shown in Table~\ref{tab:multiclass}, our method still achieves satisfactory performance on 3D detection of pedestrians and cyclists. Moreover, we also show the active maps corresponding to different filters of our D$^4$LCN in Figure~\ref{fig:filter}. Different filters on the same layer of our model use different sizes of receptive fields to handle objects of different scales, including pedestrians (small) and cars (big), as well as distant cars (big) and nearby cars (small).

\section{Conclusion}

In this paper, we propose a Depth-guided Dynamic-Depthwise-Dilated Local ConvNet (D$^4$LCN) for monocular 3D objection detection, where the convolutional kernels and their receptive fields (dilation rates) are different for different pixels and channels of different images. These kernels are generated dynamically conditioned on the depth map to compensate the limitations of 2D convolution and narrow the gap between 2D convolutions and the point cloud-based 3D operators. As a result, our D$^4$LCN can not only address the problem of the scale-sensitive and meaningless local structure of 2D convolutions, but also benefit from the high-level semantic information from RGB images. Extensive experiments show that our D$^4$LCN better captures 3D information and ranks $1^{\mathrm{st}}$ for monocular 3D object detection on the KITTI dataset at the time of submission.

\section{Acknowledgements}

We would like to thank Dr. Guorun Yang for his careful proofreading. Ping Luo is partially supported by the HKU Seed Funding for Basic Research and SenseTime's Donation for Basic Research. Zhiwu Lu is partially supported by National Natural Science Foundation of China (61976220, 61832017, and 61573363), and Beijing Outstanding Young Scientist Program (BJJWZYJH012019100020098).

\begin{appendices}
\vspace{20pt}
\noindent \textbf{\large APPENDIX}

\section{Definition of 3D Corners}
We define the eight corners of each ground truth box as follows:
\begin{small}
\begin{align}
C^{(m)} = \begin{bmatrix}
    x^{(m)} \\
    y^{(m)} \\
    1 \\
  \end{bmatrix}_{P} \hspace{-0.05in}\cdot z^{(m)}_{3D} = \begin{pmatrix} r_y \cdot \begin{bmatrix} \pm w/2 \\ \pm h/2 \\ \pm l/2 \\ 0 \end{bmatrix}_{3D} \hspace{-0.05in}+ \begin{bmatrix} x \\ y  \\ z \\ 1\end{bmatrix}_{3D} \end{pmatrix} 
\end{align}
\end{small}
\hspace{-3pt}where $m \in (int)[1,8]$ in a defined order, and $r_y$ is the egocentric rotation matrix. Note that we use allocentric pose for regression.

\section{Comparisons between Two Rotation Definitions}

\begin{figure}[t]
\begin{center}
\includegraphics[width=1\columnwidth]{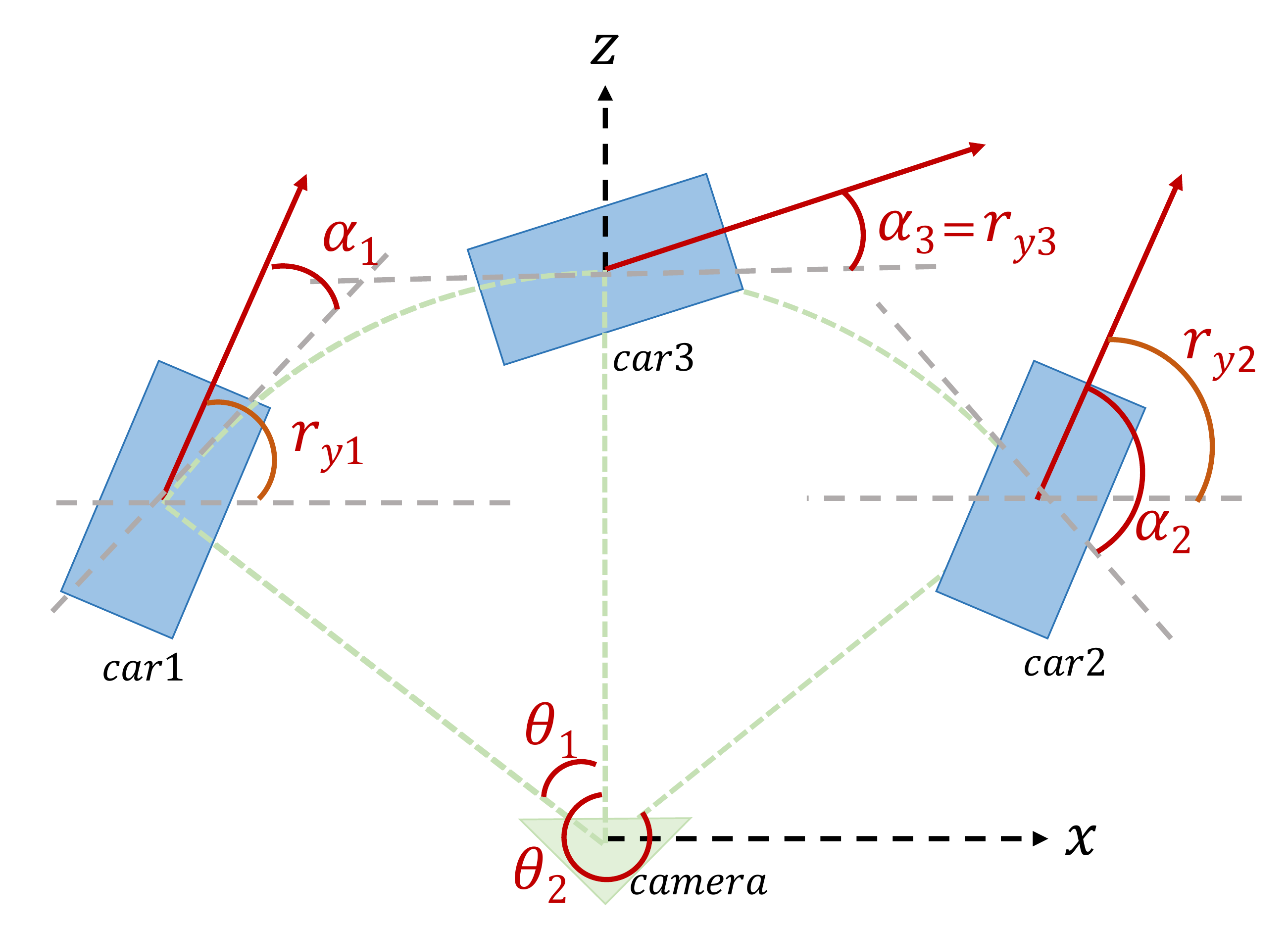}
\end{center}
   \caption{Comparisons between egocentric ($r_y$) and allocentric ($\alpha$) poses. The car1 and car2 have the same egocentric pose, but they are observed on different sides (views). We use allocentric pose to keep the same view (car1 and car3).}
\label{fig:definition}
\end{figure}

As shown in Figure~\ref{fig:definition}, while egocentric poses undergo viewpoint changes towards the camera when translated, allocentric poses always exhibit the same view, independent of the object’s location. The allocentric pose $\alpha$ and the egocentric pose $r_y$ can be converted to each other according to the viewing angle $\theta$.
\begin{align}
\alpha = r_y - \theta
\label{eq:alpha}
\end{align}

\begin{table*}[t]
\centering
\tabcolsep=0.13cm
\begin{tabular}{|c|c|c|c|c|c|c|c|c|} 
\hline
Method & Depth & CAD & Points & Freespace & Segmentation & Pretrain/MST & End-to-end \\
\hline 
\tabincell{c}{
Deep3DBox \cite{mousavian20173d}, GS3D \cite{li2019gs3d}, \\ 
MonoGRNet \cite{qin2019monogrnet}, OFTNet \cite{roddick2018orthographic} \\
FQNet \cite{liu2019deep}, SS3D \cite{jorgensen2019monocular}
}&  &  &  &  &  &  & \checkmark \\
\hline
ROI-10D \cite{manhardt2019roi} &  & \checkmark &  &  &  &  & \checkmark \\
\hline
Multi-Level Fusion \cite{xu2018multi}, \textbf{D}$\mathbf{^4}$\textbf{LCN (Ours)} & \checkmark &  &  &  &  & &  \checkmark \\
\hline
\tabincell{c}{M3D-RPN \cite{brazil2019m3d}, MONODIS \cite{simonelli2019disentangling},\\ Shift R-CNN \cite{naiden2019shift}} &  &  &  &  &  &  \checkmark & \\
\hline
\tabincell{c}{AM3D \cite{ma2019accurate} Pseudo-LiDAR \cite{wang2019pseudo},\\ Mono3D-PLiDAR\cite{weng2019monocular}, MonoPSR \cite{ku2019monocular}} & \checkmark &  &  &  &  & \checkmark &  \\
\hline
Deep-MANTA \cite{chabot2017deep} &  &  & \checkmark &  &  &\checkmark & \\
\hline
3DVP \cite{xiang2015data} &  & \checkmark &  &  &  & \checkmark & \\
\hline
Mono3D \cite{chen2016monocular,chen20173d} & \checkmark &  &  & \checkmark & \checkmark & \checkmark &  \\
\hline
Mono3D++ \cite{he2019mono3d++} & \checkmark &  & \checkmark &  &  & \checkmark &  \\
\hline
\end{tabular}
\caption{Comparisons of the labeling information and training strategies used in different monocular detection methods. Notations:
Pretrain -- pre-trained on COCO/KITTI datasets;
MST -- multi-stage training;
End-to-end -- end-to-end training;
Depth -- the depth map extracted from monocular image;
CAD -- the CAD model;
Points -- the characteristic points labeling information;
Freespace -- the free space labeling information;
Segmentation -- the segmentation labeling information;}
\label{tab:methods}
\vspace{10pt}
\end{table*}

\begin{table*}[t]
    \centering
    {\footnotesize \tabcolsep=0.17cm
    \begin{tabular}{l|c|c|c|c|c|ccc|ccc}
        \toprule
        \multirow{2}{*}{Conv Method} &
        \multirow{2}{*}{Dynamic} &
        \multirow{2}{*}{Local} &
        \multirow{2}{*}{Depth-wise} &
        \multirow{2}{*}{Shift-pooling} &
        \multirow{2}{*}{Dilated} &
        \multicolumn{3}{c|}{$\text{AP}|_{R_{11}}$} & \multicolumn{3}{c}{$\text{AP}|_{R_{40}}$}\\
         &&&&&& Easy & \textbf{Moderate} & Hard & Easy & \textbf{Moderate} & Hard \\
        \midrule
    ConvNet &&&&&& 20.66 & 15.57 & 13.41 & 17.10 & 12.09 & 09.47\\
	Depth-guided CN &\checkmark&&&&& 23.01 & 17.67 & 15.85 & 17.47 & 12.18 & 09.53 \\
	Depth-guided LCN &\checkmark&\checkmark&&&& 25.15 & 18.42 & 16.27 & 21.09 & 13.93 & 11.31 \\
	Depth-guided DLCN&\checkmark&\checkmark&\checkmark&&& 23.25 & 17.92 & 15.58 & 18.32 & 13.50 & 10.61 \\
	Depth-guided SP-DLCN &\checkmark&\checkmark&\checkmark&\checkmark&& 25.30 & 19.02 & 17.26 & 19.69 & 14.44 & 11.52 \\
	\textbf{D}$\mathbf{^4}$\textbf{LCN} &\checkmark&\checkmark&\checkmark&\checkmark&\checkmark& \textbf{26.97} & \textbf{21.71} & \textbf{18.22} & \textbf{22.32} & \textbf{16.20} & \textbf{12.30} \\
        \bottomrule
    \end{tabular}}
    \caption{Comparisons of different convolutional methods for \textit{car} 3D detection on the KITTI split1.}
    \label{tab:ablation}
\end{table*}

\section{Ablative Results for Convolutional Methods}
The Depth-guided filtering module in our D$^4$LCN model can be decomposed into basic convolutional components:
\begin{itemize}
\item Traditional Convolutional Network
\item Depth-guided ConvNet (CN)
\item Depth-guided Local CN (LCN)
\item Depth-guided Depth-wise LCN (DLCN)
\item Depth-guided DLCN with Shift-pooling (SP-DLCN)
\item D$^4$LCN (Our full model)
\end{itemize}
The ablative results for these convolutional methods are shown in Table~\ref{tab:ablation}. We can observe that:
(1) Using the depth map to guide the convolution of each pixel brings a considerable improvement.
(2) Depth-wise convolution with shift-pooling operator not only has fewer parameters (Section 3.2 of our main paper) but also gets better performance than the standard convolution.
(3) The main improvement comes from our adaptive dilated convolution, which allows each channel of the feature map to have different receptive fields.

\section{Comparisons of Labeling Information and Training Strategies}

We compare the labeling information and training strategies used in different monocular detection methods, as shown in Table~\ref{tab:methods}.

It can be seen that: 
(1) our model outperforms all existing methods by only using the depth map extracted from the monocular image. (2) our model can be trained in an end-to-end manner.

\section{Distributions of Different Dilation}

\begin{figure}[h]
\begin{center}
    \includegraphics[width=1\columnwidth]{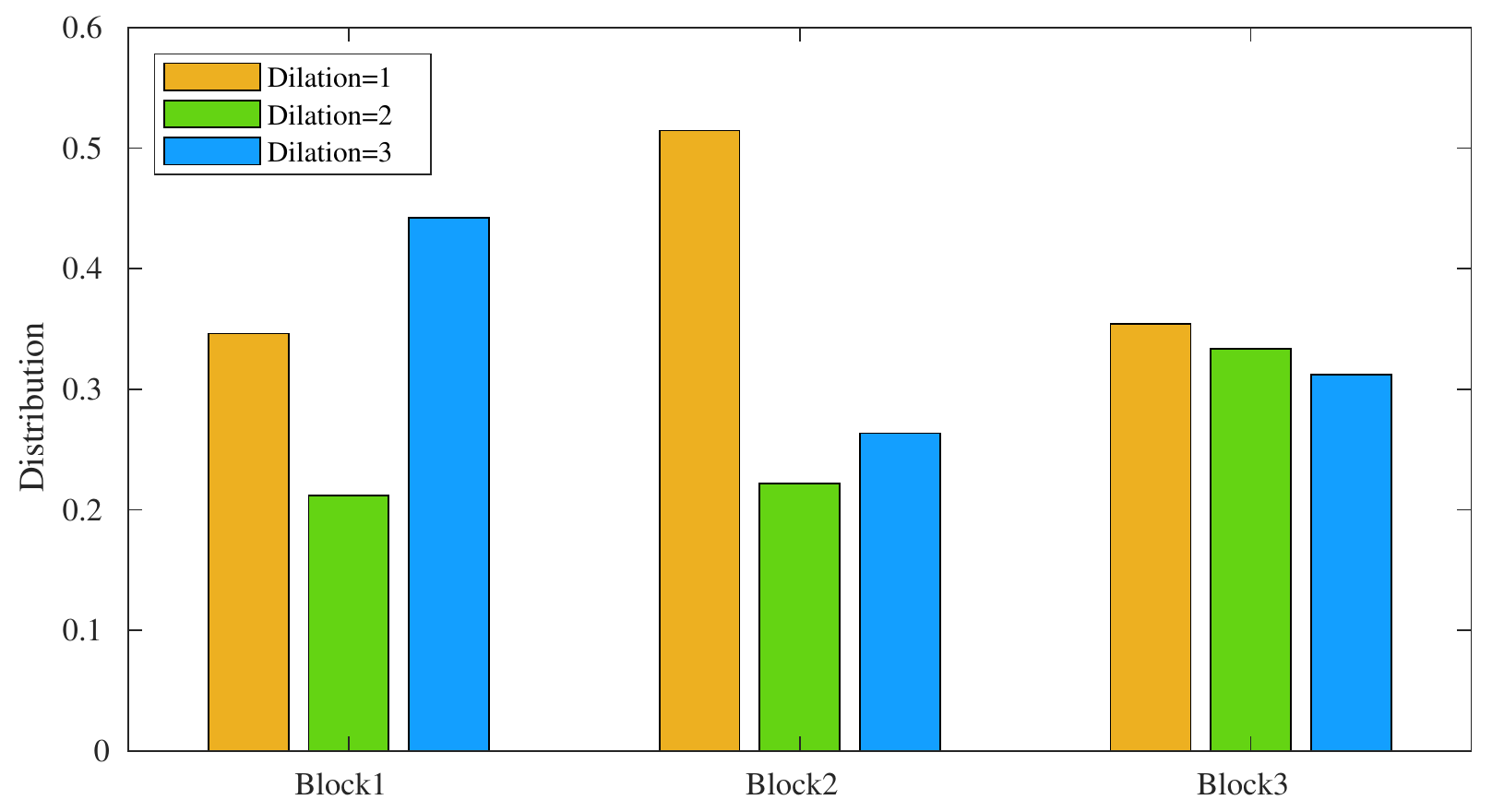}
\end{center}
   \caption{The average ratio of different channels with different dilation rates in three blocks.}
\label{fig:probability}
\end{figure}

We show the average ratio of different channels with different dilation rates in three blocks of our model over the validation set of split1 (Figure~\ref{fig:probability}). It can be seen that: (1) For the first block with insufficient receptive field, the model tends to increase the receptive field by large dilation rate, and then it uses small receptive field for the second block. (2) In the third block, the model uses three different dilation rates evenly to deal with the object detection of different scales. We also show the active maps corresponding to different filters of the third block of our D$^4$LCN in our main paper (Figure 5).

\end{appendices}

{\small
\bibliographystyle{ieee_fullname}
\bibliography{egbib}

\begin{thebibliography}{10}\itemsep=-1pt

\bibitem{brazil2019m3d}
Garrick Brazil and Xiaoming Liu.
\newblock M3d-rpn: Monocular 3d region proposal network for object detection.
\newblock In {\em ICCV}, pages 9287--9296, 2019.

\bibitem{chabot2017deep}
Florian Chabot, Mohamed Chaouch, Jaonary Rabarisoa, C{\'e}line Teuli{\`e}re,
  and Thierry Chateau.
\newblock Deep manta: A coarse-to-fine many-task network for joint 2d and 3d
  vehicle analysis from monocular image.
\newblock In {\em CVPR}, pages 2040--2049, 2017.

\bibitem{chang2018pyramid}
Jia-Ren Chang and Yong-Sheng Chen.
\newblock Pyramid stereo matching network.
\newblock In {\em CVPR}, pages 5410--5418, 2018.

\bibitem{chen2016monocular}
Xiaozhi Chen, Kaustav Kundu, Ziyu Zhang, Huimin Ma, Sanja Fidler, and Raquel
  Urtasun.
\newblock Monocular 3d object detection for autonomous driving.
\newblock In {\em CVPR}, pages 2147--2156, 2016.

\bibitem{chen20153d}
Xiaozhi Chen, Kaustav Kundu, Yukun Zhu, Andrew~G Berneshawi, Huimin Ma, Sanja
  Fidler, and Raquel Urtasun.
\newblock 3d object proposals for accurate object class detection.
\newblock In {\em NeurIPS}, pages 424--432, 2015.

\bibitem{chen20173d}
Xiaozhi Chen, Kaustav Kundu, Yukun Zhu, Huimin Ma, Sanja Fidler, and Raquel
  Urtasun.
\newblock 3d object proposals using stereo imagery for accurate object class
  detection.
\newblock {\em TPAMI}, 40(5):1259--1272, 2017.

\bibitem{chen2017multi}
Xiaozhi Chen, Huimin Ma, Ji Wan, Bo Li, and Tian Xia.
\newblock Multi-view 3d object detection network for autonomous driving.
\newblock In {\em CVPR}, pages 1907--1915, 2017.

\bibitem{chen2019fast}
Yilun Chen, Shu Liu, Xiaoyong Shen, and Jiaya Jia.
\newblock Fast point r-cnn.
\newblock In {\em ICCV}, 2019.

\bibitem{dai2017deformable}
Jifeng Dai, Haozhi Qi, Yuwen Xiong, Yi Li, Guodong Zhang, Han Hu, and Yichen
  Wei.
\newblock Deformable convolutional networks.
\newblock In {\em ICCV}, pages 764--773, 2017.

\bibitem{deng2009imagenet}
Jia Deng, Wei Dong, Richard Socher, Li-Jia Li, Kai Li, and Li Fei-Fei.
\newblock Imagenet: A large-scale hierarchical image database.
\newblock In {\em CVPR}, pages 248--255, 2009.

\bibitem{fu2018deep}
Huan Fu, Mingming Gong, Chaohui Wang, Kayhan Batmanghelich, and Dacheng Tao.
\newblock Deep ordinal regression network for monocular depth estimation.
\newblock In {\em CVPR}, pages 2002--2011, 2018.

\bibitem{geiger2012we}
Andreas Geiger, Philip Lenz, and Raquel Urtasun.
\newblock Are we ready for autonomous driving? the kitti vision benchmark
  suite.
\newblock In {\em CVPR}, pages 3354--3361, 2012.

\bibitem{monodepth17}
Cl{\'{e}}ment Godard, Oisin {Mac Aodha}, and Gabriel~J. Brostow.
\newblock Unsupervised monocular depth estimation with left-right consistency.
\newblock In {\em CVPR}, 2017.

\bibitem{ha2016hypernetworks}
David Ha, Andrew Dai, and Quoc~V Le.
\newblock Hypernetworks.
\newblock {\em arXiv preprint arXiv:1609.09106}, 2016.

\bibitem{he2010guided}
Kaiming He, Jian Sun, and Xiaoou Tang.
\newblock Guided image filtering.
\newblock In {\em ECCV}, pages 1--14, 2010.

\bibitem{he2016deep}
Kaiming He, Xiangyu Zhang, Shaoqing Ren, and Jian Sun.
\newblock Deep residual learning for image recognition.
\newblock In {\em CVPR}, pages 770--778, 2016.

\bibitem{he2019mono3d++}
Tong He and Stefano Soatto.
\newblock Mono3d++: Monocular 3d vehicle detection with two-scale 3d hypotheses
  and task priors.
\newblock {\em arXiv preprint arXiv:1901.03446}, 2019.

\bibitem{howard2017mobilenets}
Andrew~G Howard, Menglong Zhu, Bo Chen, Dmitry Kalenichenko, Weijun Wang,
  Tobias Weyand, Marco Andreetto, and Hartwig Adam.
\newblock Mobilenets: Efficient convolutional neural networks for mobile vision
  applications.
\newblock {\em arXiv preprint arXiv:1704.04861}, 2017.

\bibitem{hu2019joint}
Hou-Ning Hu, Qi-Zhi Cai, Dequan Wang, Ji Lin, Min Sun, Philipp Krahenbuhl,
  Trevor Darrell, and Fisher Yu.
\newblock Joint monocular 3d vehicle detection and tracking.
\newblock In {\em ICCV}, pages 5390--5399, 2019.

\bibitem{jia2016dynamic}
Xu Jia, Bert De~Brabandere, Tinne Tuytelaars, and Luc~V Gool.
\newblock Dynamic filter networks.
\newblock In {\em NeurIPS}, pages 667--675, 2016.

\bibitem{jorgensen2019monocular}
Eskil J{\"o}rgensen, Christopher Zach, and Fredrik Kahl.
\newblock Monocular 3d object detection and box fitting trained end-to-end
  using intersection-over-union loss.
\newblock {\em arXiv preprint arXiv:1906.08070}, 2019.

\bibitem{ku2018joint}
Jason Ku, Melissa Mozifian, Jungwook Lee, Ali Harakeh, and Steven~L Waslander.
\newblock Joint 3d proposal generation and object detection from view
  aggregation.
\newblock In {\em 2018 IEEE/RSJ International Conference on Intelligent Robots
  and Systems (IROS)}, pages 1--8, 2018.

\bibitem{ku2019monocular}
Jason Ku, Alex~D Pon, and Steven~L Waslander.
\newblock Monocular 3d object detection leveraging accurate proposals and shape
  reconstruction.
\newblock In {\em CVPR}, pages 11867--11876, 2019.

\bibitem{kundu20183d}
Abhijit Kundu, Yin Li, and James~M Rehg.
\newblock 3d-rcnn: Instance-level 3d object reconstruction via
  render-and-compare.
\newblock In {\em CVPR}, pages 3559--3568, 2018.

\bibitem{lang2019pointpillars}
Alex~H Lang, Sourabh Vora, Holger Caesar, Lubing Zhou, Jiong Yang, and Oscar
  Beijbom.
\newblock Pointpillars: Fast encoders for object detection from point clouds.
\newblock In {\em CVPR}, pages 12697--12705, 2019.

\bibitem{li2019gs3d}
Buyu Li, Wanli Ouyang, Lu Sheng, Xingyu Zeng, and Xiaogang Wang.
\newblock Gs3d: An efficient 3d object detection framework for autonomous
  driving.
\newblock In {\em CVPR}, pages 1019--1028, 2019.

\bibitem{li2019scale}
Yanghao Li, Yuntao Chen, Naiyan Wang, and Zhaoxiang Zhang.
\newblock Scale-aware trident networks for object detection.
\newblock In {\em ICCV}, 2019.

\bibitem{liang2019multi}
Ming Liang, Bin Yang, Yun Chen, Rui Hu, and Raquel Urtasun.
\newblock Multi-task multi-sensor fusion for 3d object detection.
\newblock In {\em CVPR}, pages 7345--7353, 2019.

\bibitem{liang2018deep}
Ming Liang, Bin Yang, Shenlong Wang, and Raquel Urtasun.
\newblock Deep continuous fusion for multi-sensor 3d object detection.
\newblock In {\em ECCV}, pages 641--656, 2018.

\bibitem{lin2017focal}
Tsung-Yi Lin, Priya Goyal, Ross Girshick, Kaiming He, and Piotr Doll{\'a}r.
\newblock Focal loss for dense object detection.
\newblock In {\em ICCV}, pages 2980--2988, 2017.

\bibitem{liu2019deep}
Lijie Liu, Jiwen Lu, Chunjing Xu, Qi Tian, and Jie Zhou.
\newblock Deep fitting degree scoring network for monocular 3d object
  detection.
\newblock In {\em CVPR}, pages 1057--1066, 2019.

\bibitem{liu2016ssd}
Wei Liu, Dragomir Anguelov, Dumitru Erhan, Christian Szegedy, Scott Reed,
  Cheng-Yang Fu, and Alexander~C Berg.
\newblock Ssd: Single shot multibox detector.
\newblock In {\em ECCV}, pages 21--37, 2016.

\bibitem{ma2019accurate}
Xinzhu Ma, Zhihui Wang, Haojie Li, Pengbo Zhang, Wanli Ouyang, and Xin Fan.
\newblock Accurate monocular 3d object detection via color-embedded 3d
  reconstruction for autonomous driving.
\newblock In {\em ICCV}, pages 6851--6860, 2019.

\bibitem{manhardt2019roi}
Fabian Manhardt, Wadim Kehl, and Adrien Gaidon.
\newblock Roi-10d: Monocular lifting of 2d detection to 6d pose and metric
  shape.
\newblock In {\em CVPR}, pages 2069--2078, 2019.

\bibitem{mayer2016large}
Nikolaus Mayer, Eddy Ilg, Philip Hausser, Philipp Fischer, Daniel Cremers,
  Alexey Dosovitskiy, and Thomas Brox.
\newblock A large dataset to train convolutional networks for disparity,
  optical flow, and scene flow estimation.
\newblock In {\em CVPR}, pages 4040--4048, 2016.

\bibitem{mousavian20173d}
Arsalan Mousavian, Dragomir Anguelov, John Flynn, and Jana Kosecka.
\newblock 3d bounding box estimation using deep learning and geometry.
\newblock In {\em CVPR}, pages 7074--7082, 2017.

\bibitem{naiden2019shift}
Andretti Naiden, Vlad Paunescu, Gyeongmo Kim, ByeongMoon Jeon, and Marius
  Leordeanu.
\newblock Shift r-cnn: Deep monocular 3d object detection with closed-form
  geometric constraints.
\newblock {\em arXiv preprint arXiv:1905.09970}, 2019.

\bibitem{qi2018frustum}
Charles~R Qi, Wei Liu, Chenxia Wu, Hao Su, and Leonidas~J Guibas.
\newblock Frustum pointnets for 3d object detection from rgb-d data.
\newblock In {\em CVPR}, pages 918--927, 2018.

\bibitem{qi2017pointnet}
Charles~R Qi, Hao Su, Kaichun Mo, and Leonidas~J Guibas.
\newblock Pointnet: Deep learning on point sets for 3d classification and
  segmentation.
\newblock In {\em CVPR}, pages 652--660, 2017.

\bibitem{qi2017pointnet++}
Charles~Ruizhongtai Qi, Li Yi, Hao Su, and Leonidas~J Guibas.
\newblock Pointnet++: Deep hierarchical feature learning on point sets in a
  metric space.
\newblock In {\em NeurIPS}, pages 5099--5108, 2017.

\bibitem{qin2019monogrnet}
Zengyi Qin, Jinglu Wang, and Yan Lu.
\newblock Monogrnet: A geometric reasoning network for monocular 3d object
  localization.
\newblock In {\em AAAI}, volume~33, pages 8851--8858, 2019.

\bibitem{redmon2016you}
Joseph Redmon, Santosh Divvala, Ross Girshick, and Ali Farhadi.
\newblock You only look once: Unified, real-time object detection.
\newblock In {\em CVPR}, pages 779--788, 2016.

\bibitem{roddick2018orthographic}
Thomas Roddick, Alex Kendall, and Roberto Cipolla.
\newblock Orthographic feature transform for monocular 3d object detection.
\newblock {\em arXiv preprint arXiv:1811.08188}, 2018.

\bibitem{shi2019pointrcnn}
Shaoshuai Shi, Xiaogang Wang, and Hongsheng Li.
\newblock Pointrcnn: 3d object proposal generation and detection from point
  cloud.
\newblock In {\em CVPR}, pages 770--779, 2019.

\bibitem{simonelli2019disentangling}
Andrea Simonelli, Samuel Rota~Rota Bul{\`o}, Lorenzo Porzi, Manuel
  L{\'o}pez-Antequera, and Peter Kontschieder.
\newblock Disentangling monocular 3d object detection.
\newblock {\em arXiv preprint arXiv:1905.12365}, 2019.

\bibitem{tang2019learning}
Jie Tang, Fei-Peng Tian, Wei Feng, Jian Li, and Ping Tan.
\newblock Learning guided convolutional network for depth completion.
\newblock {\em arXiv preprint arXiv:1908.01238}, 2019.

\bibitem{wang2019voxel}
Bei Wang, Jianping An, and Jiayan Cao.
\newblock Voxel-fpn: multi-scale voxel feature aggregation in 3d object
  detection from point clouds.
\newblock {\em arXiv preprint arXiv:1907.05286}, 2019.

\bibitem{wang2019pseudo}
Yan Wang, Wei-Lun Chao, Divyansh Garg, Bharath Hariharan, Mark Campbell, and
  Kilian~Q Weinberger.
\newblock Pseudo-lidar from visual depth estimation: Bridging the gap in 3d
  object detection for autonomous driving.
\newblock In {\em CVPR}, pages 8445--8453, 2019.

\bibitem{wang2019frustum}
Zhixin Wang and Kui Jia.
\newblock Frustum convnet: Sliding frustums to aggregate local point-wise
  features for amodal 3d object detection.
\newblock {\em arXiv preprint arXiv:1903.01864}, 2019.

\bibitem{weng2019monocular}
Xinshuo Weng and Kris Kitani.
\newblock Monocular 3d object detection with pseudo-lidar point cloud.
\newblock {\em arXiv preprint arXiv:1903.09847}, 2019.

\bibitem{wu2018shift}
Bichen Wu, Alvin Wan, Xiangyu Yue, Peter Jin, Sicheng Zhao, Noah Golmant, Amir
  Gholaminejad, Joseph Gonzalez, and Kurt Keutzer.
\newblock Shift: A zero flop, zero parameter alternative to spatial
  convolutions.
\newblock In {\em CVPR}, pages 9127--9135, 2018.

\bibitem{wu2018dynamic}
Jialin Wu, Dai Li, Yu Yang, Chandrajit Bajaj, and Xiangyang Ji.
\newblock Dynamic filtering with large sampling field for convnets.
\newblock In {\em ECCV}, pages 185--200, 2018.

\bibitem{xiang2015data}
Yu Xiang, Wongun Choi, Yuanqing Lin, and Silvio Savarese.
\newblock Data-driven 3d voxel patterns for object category recognition.
\newblock In {\em CVPR}, pages 1903--1911, 2015.

\bibitem{xu2018multi}
Bin Xu and Zhenzhong Chen.
\newblock Multi-level fusion based 3d object detection from monocular images.
\newblock In {\em CVPR}, pages 2345--2353, 2018.

\bibitem{yan2018second}
Yan Yan, Yuxing Mao, and Bo Li.
\newblock Second: Sparsely embedded convolutional detection.
\newblock {\em Sensors}, 18(10):3337, 2018.

\bibitem{yu2015multi}
Fisher Yu and Vladlen Koltun.
\newblock Multi-scale context aggregation by dilated convolutions.
\newblock {\em arXiv preprint arXiv:1511.07122}, 2015.

\bibitem{zeeshan2014cars}
Muhammad Zeeshan~Zia, Michael Stark, and Konrad Schindler.
\newblock Are cars just 3d boxes?-jointly estimating the 3d shape of multiple
  objects.
\newblock In {\em CVPR}, pages 3678--3685, 2014.

\bibitem{zhang2018shufflenet}
Xiangyu Zhang, Xinyu Zhou, Mengxiao Lin, and Jian Sun.
\newblock Shufflenet: An extremely efficient convolutional neural network for
  mobile devices.
\newblock In {\em CVPR}, pages 6848--6856, 2018.

\bibitem{zhou2018voxelnet}
Yin Zhou and Oncel Tuzel.
\newblock Voxelnet: End-to-end learning for point cloud based 3d object
  detection.
\newblock In {\em CVPR}, pages 4490--4499, 2018.

\end{thebibliography}
}

\end{document}